\documentclass{article}

\usepackage{microtype}
\usepackage{graphicx}
\usepackage{subfigure}
\usepackage{booktabs} 

\usepackage{hyperref}

\usepackage[accepted]{icml2025}

\usepackage{amsmath}
\usepackage{amssymb}
\usepackage{enumitem}
\usepackage{mathtools}
\usepackage{amsthm}
\usepackage{float}

\usepackage[capitalize,noabbrev]{cleveref}

\theoremstyle{plain}

\theoremstyle{definition}

\theoremstyle{remark}

\usepackage[textsize=tiny]{todonotes}

\icmltitlerunning{Manimator: Transforming Research Papers into Visual Explanations}

\begin{document}

\twocolumn[
\icmltitle{Manimator: Transforming Research Papers and Mathematical \\ Concepts into Visual Explanations}



\icmlsetsymbol{equal}{*}

\begin{icmlauthorlist}
\icmlauthor{Samarth P}{equal,yyy}
\icmlauthor{Vyoman Jain}{equal,yyy}
\icmlauthor{Shiva Golugula}{equal,yyy}
\icmlauthor{Motamarri Sai Sathvik}{equal,yyy}
\end{icmlauthorlist}

\icmlaffiliation{yyy}{Department of Computer Science and Engineering, PES University, Bengaluru, India}

\icmlcorrespondingauthor{Samarth P}{pes2ug22cs495@pesu.pesu.edu}
\icmlcorrespondingauthor{Vyoman Jain}{pes2ug22cs672@pesu.pesu.edu}
\icmlcorrespondingauthor{Shiva Golugula}{pes2ug22cs525@pesu.pesu.edu}
\icmlcorrespondingauthor{Motamarri Sai Sathvik}{pes2ug22cs321@pesu.pesu.edu}


\icmlkeywords{Machine Learning, ICML}

\vskip 0.3in
 ]



\printAffiliationsAndNotice{\icmlEqualContribution} 

\begin{abstract}
Understanding complex scientific and mathematical concepts, particularly those presented in dense research papers, poses a significant challenge for learners. Dynamic visualizations can greatly enhance comprehension, but creating them manually is time-consuming and requires specialized knowledge and skills. We introduce manimator, an open-source system that leverages Large Language Models to transform research papers and natural language prompts into explanatory animations using the Manim engine. Manimator employs a pipeline where an LLM interprets the input text or research paper PDF to generate a structured scene description outlining key concepts, mathematical formulas, and visual elements and another LLM translates this description into executable Manim Python code. We discuss its potential as an educational tool for rapidly creating engaging visual explanations for complex STEM topics, democratizing the creation of high-quality educational content.
\end{abstract}

\section{Introduction}
\label{Introduction}
The ability to grasp complex scientific and mathematical concepts is fundamental to STEM education. However, traditional learning materials, such as textbooks and research papers, often present information in dense, static formats. Dynamic visualizations have proven effective in improving understanding and engagement, making abstract concepts more concrete and intuitive \citep{show}.

The Manim animation engine \citep{manim:2024} has demonstrated the power of code-driven animations for explaining intricate topics. Despite the potential, creating such high-quality animations requires significant effort, programming expertise (Python), and domain knowledge. This bottleneck limits the widespread availability of custom visualizations tailored to specific concepts or research findings. However, recent advancements in Large Language Models (LLMs) have shown remarkable capabilities in text understanding, structured data generation, and code synthesis \citep{nijkamp:2023, chen:2023a}.

In this paper, we present manimator, a system that harnesses LLMs to bridge the gap between complex textual information and engaging visual explanations. Manimator takes either a natural language prompt describing a concept or a full research paper (via PDF upload or using its arXiv ID) as input.

It employs a multi-stage pipeline:
\begin{enumerate}[nosep]
    \item \textbf{Scene Understanding and Planning:} An LLM analyzes the input to extract key concepts, relevant formulas, and potential visual representations, structuring them into a detailed ``scene description.''
    \item \textbf{Code Generation:} A code-specialized LLM translates the structured scene description into Python code compatible with the Manim engine.
    \item \textbf{Rendering:} The generated Manim code is executed to produce a video animation.
\end{enumerate}

Manimator aims to empower educators, students, and researchers to quickly visualize complex ideas, thereby enhancing learning and scientific communication. We plan to open-source the tool and provide public access through a Gradio interface and API.

\section{Related Works}
\label{Related Works}

Our work on manimator intersects with several research areas, including visualization in STEM education, code-driven animation tools, the application of Large Language Models (LLMs) for understanding and generation, and automated multimodal content creation.

\begin{figure*}[h]
  \centering
  \includegraphics[width=0.99\textwidth]{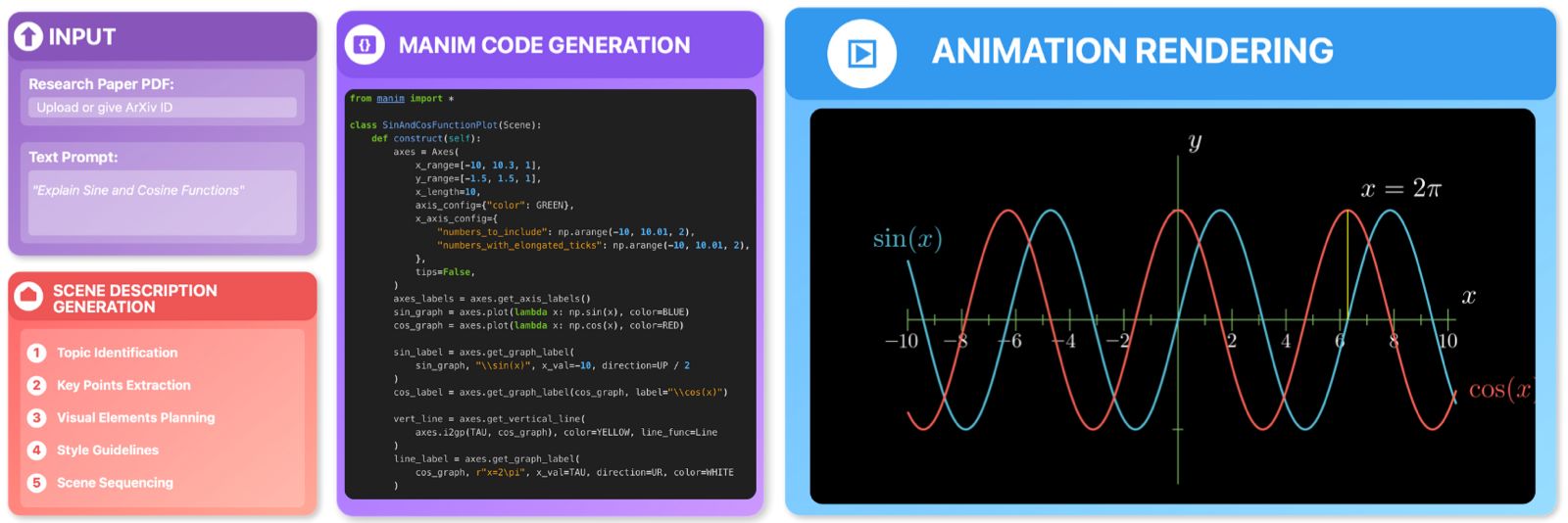}
  \caption{The Manimator Workflow: User provides input via a natural language prompt, PDF upload, or arXiv ID. The pipeline has three stages: (1) an LLM generates a structured Markdown scene description from input using prompts and examples; (2) a code-focused LLM converts this into Manim Python code; (3) the code is rendered into an animation using the Manim engine.}
  \label{fig:architecture}
\end{figure*}

\subsection{Visualization in STEM Education and Communication}

The efficacy of visualization, particularly dynamic visualization, in enhancing the understanding of complex scientific and mathematical concepts is well-established \citep{brown:1985, heer:2007}. Visual representations can make abstract ideas more concrete, reveal hidden patterns, and improve learner engagement \citep{amini:2018}. Static visualizations are common in textbooks and papers, but dynamic animations offer significant advantages for illustrating processes, algorithms, and time-varying phenomena \citep{heer:2007}. Research has explored creating visualizations for specific domains, such as algorithm and data structure education \citep{markovic:2024}. Tools like manimator aim to build upon these principles by significantly lowering the barrier to creating dynamic visualizations, making this powerful pedagogical approach more accessible. \citet{ku:2025} also highlight the benefit of visual explanations for theorem understanding, noting that they can expose reasoning errors more clearly than text alone.

\subsection{Code-Driven Animation and Manim}

The Manim library \citep{manim:2024} stands out as a powerful tool for creating precise, programmatically generated mathematical animations. Its code-driven nature allows for fine-grained control over visual elements and complex transformations, making it ideal for explaining intricate STEM topics \citep{markovic:2024}. However, creating animations with Manim requires significant programming expertise (Python) and considerable manual effort, representing a bottleneck for educators and researchers who may lack the time or specific skills \citep{markovic:2024, helbling:2023}. Prior work like \citet{helbling:2023} aimed to simplify Manim usage for specific domains like machine learning architectures by providing higher-level abstractions, but still required manual coding within its framework. Manimator directly addresses the creation bottleneck by automating the Manim code generation process itself using LLMs.

\subsection{Large Language Models for Understanding and Code Generation}

Recent advancements in Large Language Models (LLMs) like GPT-4 \citep{openai:2023}, Claude \citep{anthropic:2024}, Gemini \citep{gemini:2024}, DeepSeek \cite{deepseekai:2025} and others have demonstrated remarkable capabilities in natural language understanding, information extraction, structured data generation, and code synthesis \citep{nijkamp:2023, chen:2023a}. LLMs can process complex textual inputs, including dense academic language found in research papers, to identify key concepts and relationships. Furthermore, their ability to generate code in various programming languages, including Python, has been extensively documented and applied in tools aiming to assist or automate software development \citep{jimenez:2024}. Several studies have explored using LLMs to generate code for specific visualization tasks, such as creating static plots from natural language descriptions or data \citep{galimzyanov:2024, yang:2024b, goswami:2025}. Manimator leverages these dual capabilities: using LLMs first for deep understanding and planning (parsing text/PDFs into a scene description) and then for code generation (translating the plan into executable Manim code).

\subsection{Automated Generation of Multimodal Explanations}

Bridging the gap between textual information and visual explanations automatically is an emerging research direction. Most current approaches focus on text-to-image or text-to-static-diagram generation. Generating dynamic, structured animations for complex explanatory purposes presents unique challenges involving temporal consistency, accurate representation of symbolic information (like formulas), and pedagogical structuring. The most closely related work is TheoremExplainAgent \citep{ku:2025}, which introduces an agentic approach using LLMs to generate Manim explanation videos specifically for mathematical theorems. It employs planning and coding agents to produce structured videos, demonstrating the potential of LLMs in this domain. While manimator shares the goal of automated Manim video generation, it positions itself as a more general tool applicable to a wider range of STEM concepts, directly accepting research papers (via PDF/arXiv ID) or natural language prompts as input and utilizing a pipeline potentially suitable for various complex topics beyond formal theorems.

\begin{table*}[ht]
\centering
\small
\begin{tabular}{@{}lcccccc@{}}
\toprule
\textbf{Agent/Model} & \textbf{Accuracy} & \textbf{Visual} & \textbf{Logical} & \textbf{Element} & \textbf{Visual} & \textbf{Overall} \\
 & \textbf{\& Depth} & \textbf{Relevance} & \textbf{Flow} & \textbf{Layout} & \textbf{Consistency} & \textbf{Score} \\
\midrule
\textbf{manimator} (DeepSeek V3) & \textbf{0.77} & \textbf{0.899} & 0.88 & \textbf{0.853} & 0.852 & \textbf{0.845} \\
Claude 3.5-Sonnet  \citep{ku:2025} & 0.75 & 0.87 & 0.88 & 0.57 & \textbf{0.92} & 0.79 \\
o3-mini (medium) \citep{ku:2025} & 0.76 & 0.76 & \textbf{0.89} & 0.61 & 0.88 & 0.77 \\
\bottomrule
\end{tabular}
\caption{Comparison of Manimator's performance on TheoremExplainBench against baseline models from \citet{ku:2025}. Scores range from 0 to 1, with higher being better. Manimator's results (using DeepSeek V3) are highlighted.}
\label{tab:teb_results}
\end{table*}

\begin{table*}[ht]
\centering
\small 
\begin{tabular}{@{}lcccccc@{}} 
\toprule
\textbf{Agent/Model} & \textbf{Accuracy} & \textbf{Visual} & \textbf{Logical} & \textbf{Element} & \textbf{Visual} & \textbf{Overall} \\
 & \textbf{\& Depth} & \textbf{Relevance} & \textbf{Flow} & \textbf{Layout} & \textbf{Consistency} & \textbf{Score} \\
\midrule
\textbf{manimator} (DeepSeek V3) & 0.89 & 0.69 & 0.83 & 0.52 & 0.75 & 0.738 \\
\bottomrule
\end{tabular}
\caption{Results of human evaluation for Manimator (DeepSeek V3) based on user ratings across the same dimensions used in TheoremExplainBench. See Appendix~\ref{human_eval} for details on the evaluation setup.}
\label{tab:human_eval_results}
\end{table*}

\section{Methodology}
\label{Methodology}

Manimator employs an agentic architecture orchestrated primarily by LLMs, with the core process involving transformation of a text prompt into a video animation via intermediate structured planning and code generation stages (illustrated in Figure~\ref{fig:architecture}). Manimator accepts two primary input types: a natural language description (e.g., ``Explain the Fourier Transform'') or a PDF document provided via upload or arXiv ID, with PDF content optionally compressed and base64 encoded for multimodal LLM input.

\subsection{Stage 1: Scene Description Generation}

This stage interprets the input to create a structured scene description for the animation. Models with text and multimodal capabilities are strategically deployed based on the input type: PDF inputs leverage the multimodal capability of models with large context windows (e.g., \texttt{gemini-2.0-flash}) which can analyze and understand document content, while text prompts utilize powerful LLMs (e.g., \texttt{llama-3.3-70b} \cite{grattafiori2024llama3herdmodels}) for comprehensive scene interpretation. A detailed system prompt instructs the LLM to structure content into: Topic, Key Points (with LaTeX math like $F(x)$), Visual Elements, and Style. Few-Shot Learning examples guide the LLM towards the desired output format and detail, ensuring consistency regardless of input type. This approach allows Manimator to extract meaningful content from diverse input sources while maintaining pedagogical quality. The output is a structured Scene Description in Markdown that serves as the foundation for subsequent animation generation.

\subsection{Stage 2: Manim Code Generation}

This stage translates the Scene Description into executable Manim Python code. The Scene Description is fed to a code-focused LLM (e.g., \texttt{deepseek-v3} \cite{deepseekai:2025}), which operates under a comprehensive system prompt that positions the model as a Manim expert. This system prompt provides detailed instructions for the LLM to create educational animations by understanding topics, planning animations with logical flow, writing modular code with proper comments, and ensuring visual clarity. Few-shot examples are also included in the prompt to demonstrate proper implementation patterns and guide the model toward producing high-quality code. The prompt emphasizes important technical requirements such as avoiding element overlap, implementing proper scene cleanup, adding appropriate wait calls for pacing, and maintaining consistent animation style. The output is a fully executable Python script containing a Manim Scene class that can be directly rendered to produce the educational animation.

\subsection{Stage 3: Animation Rendering}

The generated Manim code is executed using the Manim library \citep{manim:2024} to produce the final video animation file (typically in MP4 format). This stage involves running the Python script in an environment where Manim and its dependencies (like FFmpeg) are installed.

\subsection{Model Selection and Prompt Optimization}

The performance of the manimator pipeline is critically dependent on the choice of LLMs and the design of the system prompts for each stage. We conducted extensive experimentation to identify the optimal configuration. Our process involved careful modulation of the system prompts, where we discovered that detailed, stage-wise and role-playing instructions (like ``You are an expert in creating educational animations using Manim") combined with few-shot examples yielded the most consistent and high-quality results for both scene planning and code generation (see figure~\ref{fig:system_prompts_combined}).

To identify the most effective models, we evaluated a range of state-of-the-art LLMs, including Claude 3.7 Sonnet, Llama 3.3 70B, Qwen2.5 Coder 32B, OpenAI's o3, and DeepSeek-V3. The evaluation criteria included the logical coherence of the generated scene description, the syntactic correctness and complexity of the Manim code, inference latency, and API cost. Although OpenAI o3 produced competent outputs, we found that DeepSeek-V3 offered the best price-to-performance ratio across this range of models. It consistently generated high-quality executable Manim code that adhered well to the structured plan and required minimal post-generation correction, making it our primary choice for the system. This balance of generation quality and economic viability is crucial for making manimator an accessible and scalable tool.

\section{Results and Evaluation}
\label{results}

To evaluate the quality and effectiveness of the animations generated by manimator, we primarily utilized the \textbf{TheoremExplainBench} (TEB) benchmark introduced by \citet{ku:2025}. TEB provides a standardized framework for assessing AI systems' ability to generate multimodal theorem explanations, evaluating videos across five key dimensions: Accuracy and Depth, Visual Relevance, Logical Flow, Element Layout, and Visual Consistency. An overall score, typically the geometric mean of the individual dimensions, is also computed. We evaluated manimator, configured to use the DeepSeek V3 large language model \cite{deepseekai:2025}, on TEB by comparing its performance against relevant baseline results reported in the original TEB paper \citep{ku:2025}, specifically those for Claude 3.5-Sonnet and o3-mini (medium), which were strong performers in that study. Additionally, we conducted a human evaluation using a custom dashboard (see Appendix~\ref{human_eval}), assessing the generated videos along similar quality dimensions.

The TEB results are presented in Table~\ref{tab:teb_results}. As shown, manimator demonstrates strong performance on the TEB benchmark, particularly excelling in Visual Relevance (0.899) and Logical Flow (0.880), which suggests the generated animations are both conceptually aligned with the source material and coherently structured. Manimator significantly outperforms the baseline models (Claude 3.5-Sonnet and o3-mini) in Element Layout (0.853 compared to 0.57 and 0.61, respectively), likely benefiting from the capabilities of the DeepSeek V3 model in generating spatially aware Manim code. While Claude 3.5-Sonnet achieves a slightly higher score in Visual Consistency (0.92 versus 0.852), manimator maintains strong performance across all dimensions, resulting in the highest Overall Score (0.845 compared to 0.79 and 0.77). Its Accuracy and Depth score (0.770) is comparable to the baselines. These TEB results indicate that manimator is a robust system for producing well-structured educational animations with excellent visual organization. The human evaluation results, summarized in Table~\ref{tab:human_eval_results}, provide complementary insights based on user perception, further detailed in Appendix~\ref{human_eval}.

\section{Limitations}
\label{Limitations}

The quality of the generated animations heavily depends on the capabilities of the underlying LLMs for both understanding the input content and generating correct, complex Manim code. Handling highly intricate visual details or ensuring perfect pedagogical flow for very nuanced topics remain challenges. The system currently does not incorporate iterative refinement based on user feedback on the generated video. Future work could address these limitations by incorporating feedback loops, employing more sophisticated planning agents, and fine-tuning models specifically for Manim generation.




\section{Conclusion}
\label{Conclusion}

We presented manimator, an open-source system utilizing Large Language Models to automatically generate educational animations using the Manim engine from research papers or natural language prompts. By implementing a pipeline where LLMs first interpret the input to create a structured scene description and then translate this into executable Manim code, manimator significantly lowers the barrier to creating dynamic visualizations for complex STEM topics, which traditionally requires extensive time and expertise. Our evaluations indicate that manimator produces high-quality, well-structured animations, thereby offering a valuable tool for educators, researchers, and students to enhance learning, understanding, and scientific communication by making the creation of engaging visual explanations more accessible. The system is publicly available as an open-source project and via a web interface to facilitate broader adoption and development.

\section*{Impact Statement}
\label{Impact Statement}

The primary impact of manimator lies in its potential to democratize the creation of high-quality educational content and accelerate scientific communication. By automating the difficult and time-consuming process of creating explanatory animations, our work empowers a broader audience—educators, students, and researchers who may not have the requisite programming skills or resources to produce such visualizations manually. This can lead to more engaging and effective learning experiences in STEM fields, helping to make complex, abstract concepts more intuitive and accessible to learners worldwide.

\bibliography{custom}
\bibliographystyle{icml2025}

\onecolumn 
\appendix 

\section{Appendix}
\label{Appendix}

This appendix provides supplementary materials for Manimator, including the system prompts used and visual examples of generated animations.

\subsection{System Prompts}
\label{appendix_prompts}

\begin{figure}[H]
  \centering
  \includegraphics[width=0.7\textwidth]{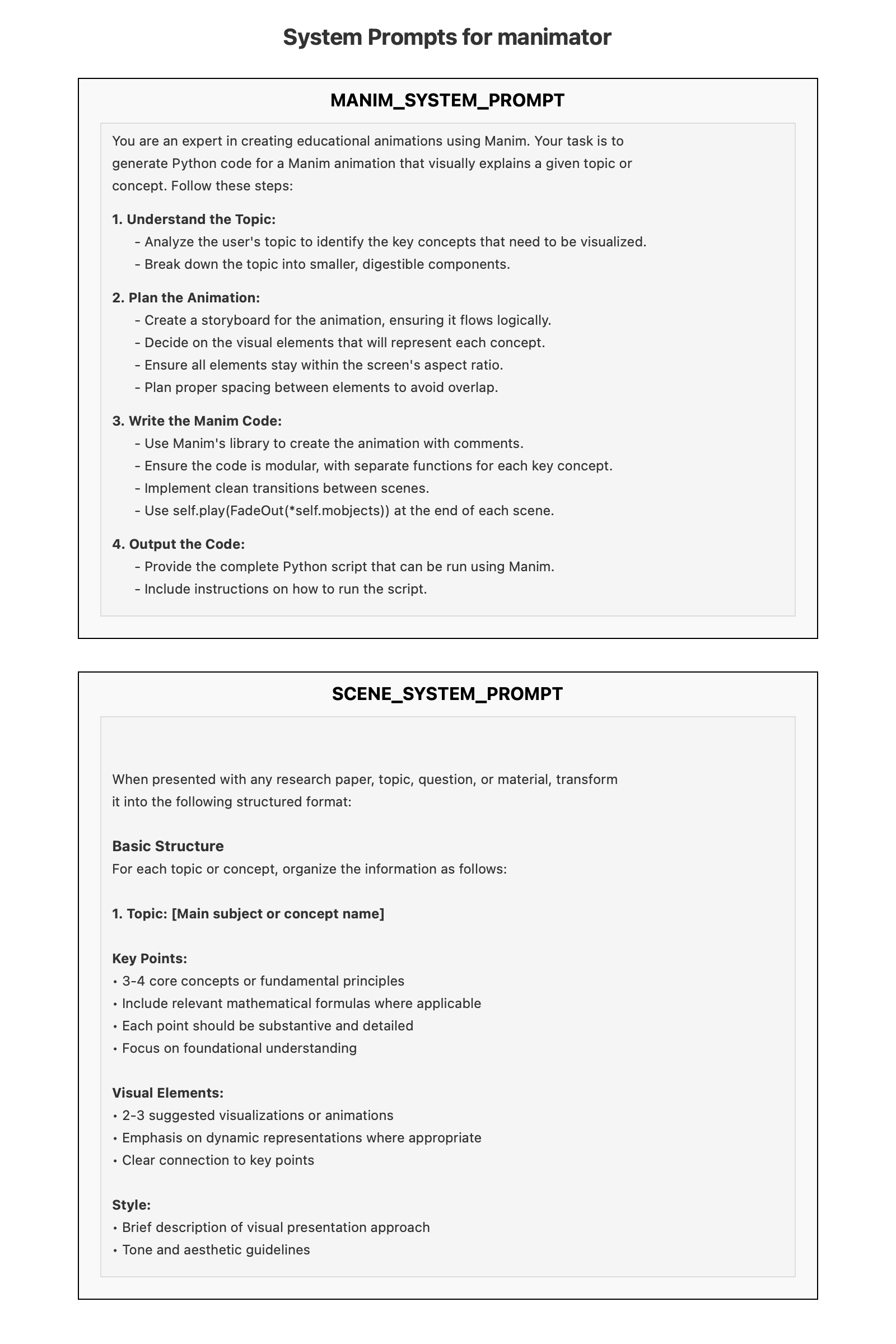}
  \caption{Combined system prompts used for Stage 1 (Scene Description Generation) and Stage 2 (Manim Code Generation).}
  \label{fig:system_prompts_combined}
\end{figure}
\clearpage

\subsection{Example Outputs}
\label{examples}
\begin{figure}[H]
  \centering
  \includegraphics[width=0.75\textwidth]{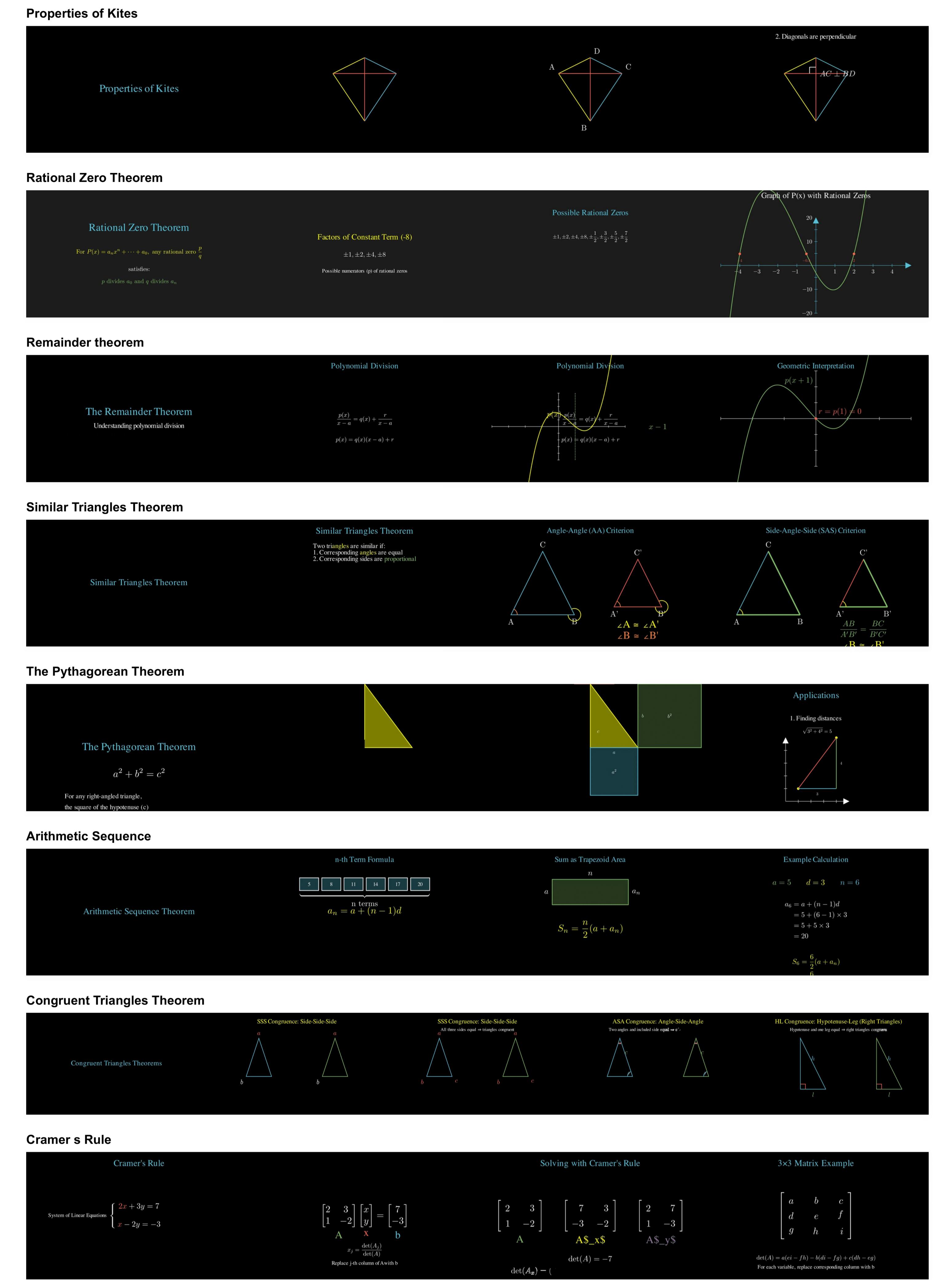}
  \caption{Examples of animations generated by Manimator.}
  \label{fig:examples}
\end{figure}

\clearpage

\subsection{Human Evaluation Dashboard}
\label{human_eval}
\begin{figure}[H]
  \centering
  \includegraphics[width=0.65\textwidth]{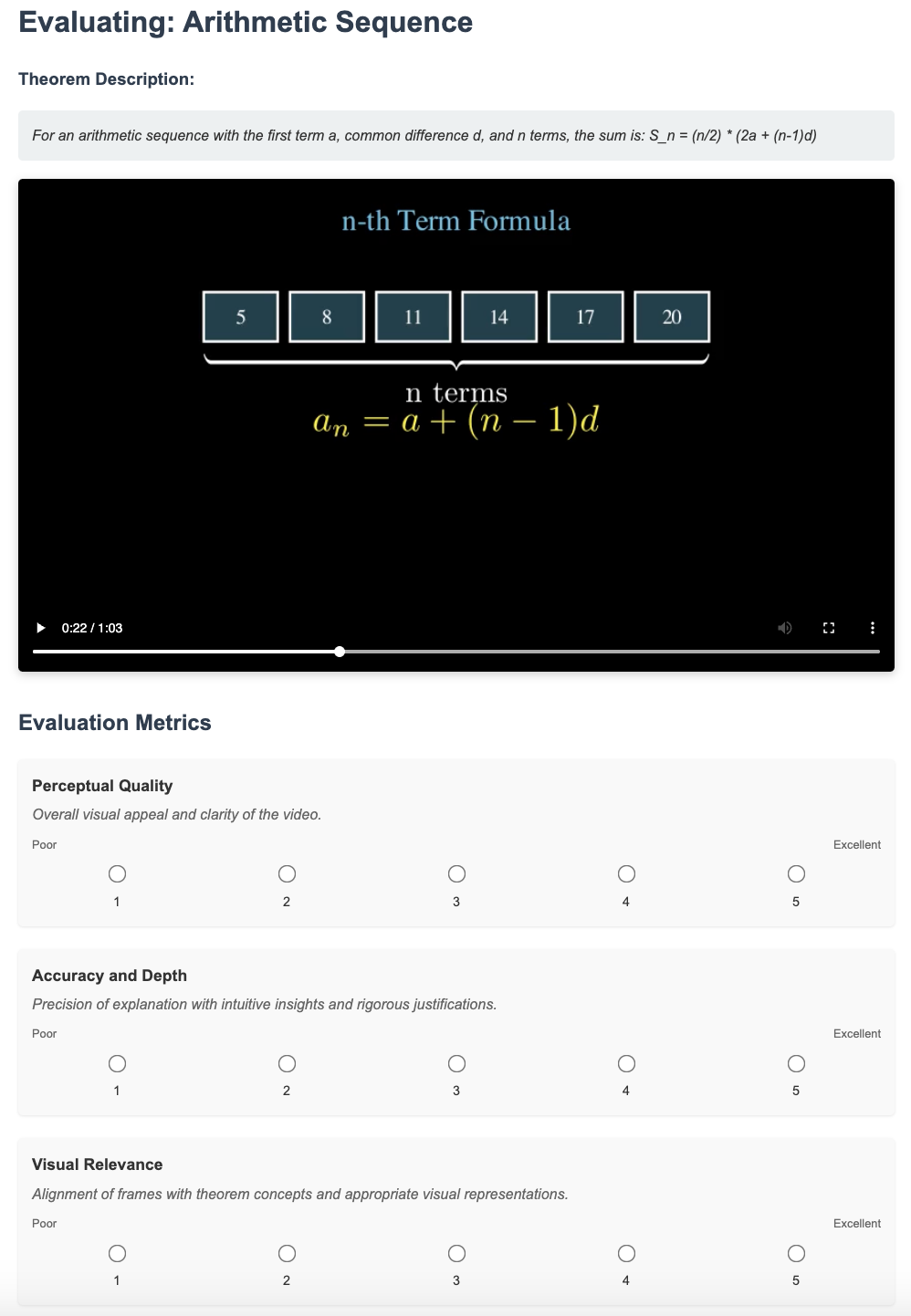}
  \caption{Interface used for collecting human evaluation ratings on generated animations.}
  \label{fig:human_eval_dashboard}
\end{figure}

To complement the standardized benchmark with direct user feedback, we recruited unpaid volunteer engineering students, who have experience with educational content like this, to participate in a human evaluation study. Participants interacted with a custom-built evaluation dashboard that presented them with animations generated by Manimator. 

For each animation, evaluators were asked to provide a rating on a scale of 1 (Poor) to 5 (Excellent) across the same five core parameters used in the TEB benchmark: Accuracy \& Depth, Visual Relevance, Logical Flow, Element Layout, and Visual Consistency. This allowed for a direct comparison between automated benchmark scores and perceived human quality. To align the 1-to-5 rating scale with the 0-to-1 scale used in TEB, we normalized the scores from our human evaluators. The final scores, representing the average normalized ratings from all participants, are summarized in Table~\ref{tab:human_eval_results}.


\end{document}